# Time-Aware Gated Recurrent Unit Networks for Road Surface Friction Prediction Using Historical Data

Ziyuan Pu [1], Zhiyong Cui[1], Shuo Wang[2], Qianmu Li[2], Yinhai Wang[3]*

[1] Smart Transportation Application and Research Laboratory, Department of Civil and Environmental Engineering, University of Washington, 101 More Hall, Seattle, WA, U.S.A.
[2] School of Computer Science and Engineering, Nanjing University of Science and Technology, 200 Xiaolingwei Street, Nanjing, Jiangsu, China.
[3] *Department of Civil and Environmental Engineering, University of Washington, 121F More Hall, Seattle, WA, U.S.A.

* corresponding author:yinhai@uw.edu

**Abstract: An accurate road surface friction prediction algorithm can enable intelligent transportation systems to share timely road surface condition to the public for increasing the safety of the road users. Previously, scholars developed multiple prediction models for forecasting road surface conditions using historical data. However, road surface condition data cannot be perfectly collected at every time stamp, e.g. the data collected by on-vehicle sensors may be influenced when vehicles cannot travel due to economic cost issue or weather issues. Such resulted missing values in the collected data can damage the effectiveness and accuracy of the existing prediction methods since they are assumed to have the input data with a fixed temporal resolution. This study proposed a road surface friction prediction model employing a Gated Recurrent Unit network-based decay mechanism (GRU-D) to handle the missing values. The evaluation results present that the proposed GRU-D networks outperform all baseline models. The impact of missing rate on predictive accuracy, learning efficiency and learned decay rate are analyzed as well. The findings can help improve the prediction accuracy and efficiency of forecasting road surface friction using historical data sets with missing values, therefore mitigating the impact of wet or icy road condition on traffic safety.**

## 1 Introduction

Road surface friction is defined as the resistance to motion between vehicle and road surface, which has strongly impact on the distance required for a vehicle to decelerate and driver's safety when a vehicle requires to break for avoiding collisions [1]. In winter season, road surface friction reduces substantially caused by dramatically decreasing temperature, which increase the risk of car accidents quite a lot [2]. FHWA reports that, in the United States, majority of traffic accidents were reported to happen during wet or icy road conditions, where 73% of accidents occurred on wet pavements, and 17% on snow or sleet [3]. In addition, existing studies indicate that the intelligent systems which have the capacity for sharing the timely road condition-related information can potentially increase the traffic safety [4]. Thus, considering the road surface friction is a directly quantitative measurement of road surface condition, an efficient and cost-effective road surface friction prediction methodology is needed for improving traffic safety.

Previously, several sensing technologies were developed for monitoring the current road surface condition parameters, e.g. DSC-111 and DST-111 sensor set [5–7], Road Condition Monitor (RCM) 411 sensor [8], and image-based sensing technology [9]. Basically, such sensing technologies can sense road surface state (dry, moist, wet, icy, snowy/frosty or slushy), pavement surface temperature, air temperature and relative humidity and etc. Previous studies evaluated the performance of the most existing sensing technologies, e.g., RCM-411 is accurate in friction level, and road surface status detection [8, 10, 11], DSC-111 can provide accurate surface state measurement, but the friction detection of DST-111 is not precise [6], etc. Some of these technologies have already been employed for real-time road monitoring implementations, e.g. Road Weather Information Station (RWIS) in US and etc.[10, 12–15].

By having such affluent data resources, scholars have developed multiple data-driven estimation and prediction models for forecasting road surface condition. Liu developed a road surface temperature prediction model based on gradient extreme learning machine boosting algorithm [16]. Solol developed a road surface temperature prediction model based on energy balance and heat conduction models [17]. In addition, some researchers developed road surface condition recognition algorithms based on computer vision technologies [18–21]. However, these studies can only estimate the current road surface condition based on environmental measurements, e.g. air temperature, etc. They are not able to predict road surface condition in the future. Considering the time-series features of road surface condition [22], multiple studies established Artificial Intelligence-based prediction model for forecasting short-term road surface condition parameters [23–25]. Basically, the inputs to these models is a set of historical road surface condition measurements with a fixed temporal resolution for each time step. The output of the forecasting model is the same road surface condition parameter in the next time step. For example, if the model targets to predict the road surface condition tomorrow, the measurements of each time step in the input data set should have 1-day time interval between each other. However, it cannot guarantee to have data points at every time stamp due to weather or cost issues, e.g., the road monitoring by on-vehicle sensors cannot be guaranteed to have a vehicle equipped with a sensor travel through the road every single day. Thus, it is highly possible to have missing values in the input data set which cannot be handled by the existing prediction models. Such missing data could damage the accuracy and effectiveness of the existing models, therefore influences the decision-making.

Based on the above considerations, the primary objective of this study is to develop a road surface friction prediction model based on time-aware recurrent gated neural networks. The RCM-411 friction sensing data were selected as the model input. To evaluate the predictive effectiveness of the proposed method, several baseline prediction and imputation models were employed for the comparison purpose. Besides the prediction performance evaluation, the impact of missing rates, the learning efficiency, and the learned decay rates are also analyzed. Findings of this study can help to improve the prediction model effectiveness by handling missing values so that mitigate the impact of road surface condition on road traffic safety and mobility.

## 2  Methodology

### 2.1  Predicting *Road Surface Friction using GRU-D Networks*

#### 2.1.1  *GRU Unit Structure*

A multivariate time series with $D$ variables of length $T$ is denoted as $X = (x_1, x_2, \cdots, x_T)^D \in \mathbb{R}^{T \times D}$, where for each time step $t \in \{1, 2, \cdots, T\}$, $x_t \in \mathbb{R}^D$ represents the measurements of all variables at time step $t$ and $x_t^d$ denotes the measurements of $d$-th variable of $x_t$. In this study, only road surface friction at time step $t$ is considered as a variable. Thus, $D$ equals 1 and a time series measurement is a $T \times 1$ vector. Then the $T$-length sequence vector is used as the input of each GRU unit. Figure 3 (a) shows the original GRU unit structure. As shown in the figure, each $j$-th GRU unit utilize a reset gate $r_t^j$ and an update gate $z_t^j$ to control the hidden state $h_t^j$ at $t$-th time step. The following equations show how the parameters are updated in each GRU unit. [26]

$$r_t = \sigma(W_r x_t + U_r h_{t-1} + b_r) \tag{1}$$

$$z_t = \sigma(W_z x_t + U_z h_{t-1} + b_z) \tag{2}$$

$$\widetilde{h_t} = tanh W x_t + U(r_t \odot h_{t-1} + b) \tag{3}$$

$$h_t = (1 - z_t) \odot h_{t-1} + z_t \widetilde{h_t} \tag{4}$$

where $W_r, W_z, W, U_r, U_z, U$ and vectors $b_r, b_z, b$ are model parameters. $\sigma$ is used for element-wise sigmoid function, and $\odot$ for element-wise multiplication. In the GRU unit update formulation, all measurements are assumed as observed values with zero missing rate. For handling missing values, the GRU-D unit will be introduced in the net section.

#### 2.1.2  *GRU-D Unit Structure*

For the time series $X$ with missing values, there are two representations of missing pattern are handled in the GRU-D unit, which are masking and time interval [27]. Masking conveys the information of which inputs are observed or missing to the model, while time interval informs GRU-D unit the time-series patterns of the input observations. A masking vector $m_t \in \{0, 1\}^D$ and the time interval vector $\delta_t^d \in \mathbb{R}$ which are shown in equations (5) and (6) are deployed for capturing the masking and time interval of the input observations.

$$m_t^d = \begin{cases} 1, & if\ x_t^d\ is\ observed \\ 0, & otherwise \end{cases} \tag{5}$$

$$\delta_t^d = \begin{cases} s_t - s_{t-1} + \delta_{t-1}^d, & t > 1, m_{t-1}^d = 0 \\ s_t - s_{t-1}, & t < 1, m_{t-1}^d = 1 \\ 0, & t = 1 \end{cases} \tag{6}$$

where $s_t \in \mathbb{R}$ denote the time stamp when the $t$-th measurement is observed and the first measurement is observed at time stamp 0. Equation (7) provides an example of the masking and time interval vectors.

$$\begin{cases} X = [x_1 \quad x_2 \quad missing \quad missing \quad x_5 \quad x_6 \quad missing \quad x_8] \\ s = [0 \quad 2 \quad 5 \quad 6 \quad 10 \quad 12 \quad 13 \quad 18] \\ M = [1 \quad 1 \quad 0 \quad 0 \quad 1 \quad 1 \quad 0 \quad 1] \\ \Delta = [0 \quad 2 \quad 3 \quad 1 \quad 8 \quad 2 \quad 1 \quad 6] \end{cases} \tag{7}$$

where $X$ is an 8-length time series measurement with three missing value at time step 3, 4 and 7. $S$ is the vector denote the time stamp when measurement $x_t$ is observed. $M$ is the vector representing the missing status of $x_t$. Then, the input data of each GRU-D unit is formed as $D = \{(X_n, s_n, M_n)_{n=1}^N\}$, where $N$ is the total number of data points in input dataset.

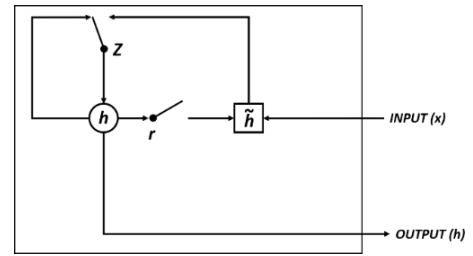

*(a) GRU Unit*

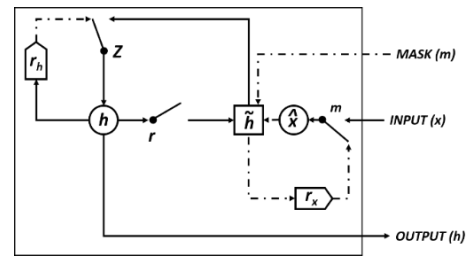

*(b) GRU-D Unit*

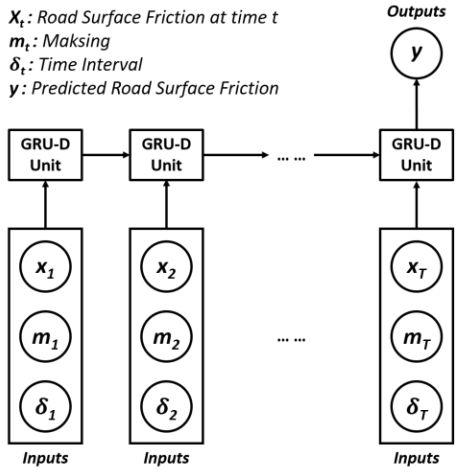

*(c)GRU-D Networks*

**Fig. 1.** *Graphical Illustrations of the Algorithm Structure. (a) Original GRU Unit Structure. (b) GRU-D Unit Structure. (c) GRU-D Networks Structure.*

As shown in figure 1 (b), GRU-D unit deploys two decay rates $\gamma_{x_t}$ and $\gamma_{h_t}$ to handle the missing values. The decay rate is calculated by equation (8).

$$\gamma_t = exp\{-max(0, W_\gamma \delta_t + b_\gamma)\} \tag{8}$$

where $W_\gamma$ and $b_\gamma$ are model parameters which are trained jointly with all the other parameters of the GRU-D unit. The decay rates are learned from the training data, and the exponentiated negative rectifier is deployed to keep the decay rate in a range between 0 and 1. Then, the input and hidden state are calculated based on equation (9) and (10).

$$\hat{x}_t^d = m_t^d x_t^d + (1 - m_t^d)(\gamma_{x_t}^d x_{t'}^d + (1 - \gamma_{x_t}^d)\tilde{x}^d) \tag{9}$$

$$\hat{h}_{t-1} = \gamma_{h_t} \odot h_{t-1} \tag{10}$$

where $x_{t'}^d$ is the last observation of the current time step, and $\tilde{x}^d$ is the empirical mean of training dataset. $h_{t-1}$ is the hidden state of the last GRU-D unit. Then, the update functions of GRU-D unit would be shown in equation (11) through (14).

$$r_t = \sigma(W_r \hat{x}_t + U_r \hat{h}_{t-1} + V_r m_t + b_r) \tag{11}$$

$$z_t = \sigma(W_z \hat{x}_t + U_z \hat{h}_{t-1} + V_z m_t + b_z) \tag{12}$$

$$\tilde{h}_t = tanh(W\hat{x}_t + U(r_t \odot \hat{h}_{t-1}) + Vm_t + b) \tag{13}$$

$$h_t = (1 - z_t) \odot \hat{h}_{t-1} + z_t \odot \tilde{h}_t \tag{14}$$

where $x_t$ and $h_{t-1}$ are substituted by $\hat{x}_t$ and $\hat{h}_{t-1}$ from equation (9) and (10). The masking vector $m_t$ are integrated with the model, and $V_r, V_z$ and $V$ are new parameters for the masking vector. In this paper, a GRU-D networks is formed by t-length GRU-D unit for the road surface friction prediction task, which is shown in figure 1 (c).

## 2.2 Predictive Performance Evaluation

### 2.2.1 Baseline Missing Data Imputation Methods

In order to evaluate the performance of GRU-D networks for road surface friction prediction, three simple methods are used to handle the missing data which are referred as Average, Last, and Simple. Equations (15) through (17) present how they handle missing data.

$$x_t^d \leftarrow m_t^d x_t^d + (1 - m_t^d)\tilde{x}^d \tag{15}$$

$$x_t^d \leftarrow m_t^d x_t^d + (1 - m_t^d)x_{t'}^d \tag{16}$$

$$x_t^d \leftarrow m_t^d x_t^d + (1 - m_t^d)(\hat{\delta}_t^d x_{t'}^d + (1 - \hat{\delta}_t^d)\tilde{x}^d) \tag{17}$$

where $\tilde{x}^d = \sum_{n=1}^{N}\sum_{t=1}^{T_n} m_{t,n}^d x_{t,n}^d / \sum_{n=1}^{N}\sum_{t=1}^{T_n} m_{t,n}^d$ , which is calculated using the training dataset and used for both training and testing dataset. $x_{t'}^d$ is the last measurements of the current time step $t$, and $\hat{\delta}_t^d$ is the normalized time interval at time step $t$.

### 2.2.2 Baseline Prediction Models

The performance of a GRU-D networks in road surface friction prediction is compared to that of many classical baseline models for short-term prediction. Typically, ARIMA, Support Vector Regression (SVR), Random forest (RF), Kalman filter, tree-based model, Feed-Forward NN and LSTM NN were used for addressing short-term prediction problems[28]–[31], e.g. traffic speed and travel time prediction [32]–[34]. However, several time-series prediction models were demonstrated that the predictive performance is not as accurate as others, e.g. ARIMA and Kalman filter. Therefore, based on previous research results, SVR, RF, Feed-Forward NN and LSTM NN were selected for comparing the performance of road surface friction prediction with the proposed GRU-D NN model in this study. Among these models, Feed-Forward NN, which is also called Multilayer Perceptron, and LSTM NN are popular for precise performance in short-term prediction [31], [35]. RF and SVR are also well-known models for efficient predictive performance [29], [36]. For the parameters of model development, Radial Basis Function (RBF) kernel is deployed in SVR model. 10 trees were built, and there was no predefined limitation for maximum depth of the trees for the RF model. The Feed-Forward NN and LSTM NN were composed of 2 hidden layers.

### 2.2.3 Evaluation Measurements

Mean Absolute Error (MAE), Mean Square Error (MSE) and Mean Absolute Percentage Error (MAPE) are used as the measurements of predictive performance. The following equations present the measurement formulation.

$$MAE = \frac{\sum_{i=1}^{N}|Y_i - \hat{Y}_i|}{N} \tag{18}$$

$$MSE = \frac{\sum_{i=1}^{N}(Y_i - \hat{Y}_i)^2}{N} \tag{19}$$

$$MAPE = \frac{100\%}{N}\sum_{i=1}^{N}\left|\frac{Y_i - \hat{Y}_i}{Y_i}\right| \tag{20}$$

where $N$ is the total number of samples in testing date set, $Y_i$ is the ground truth of the road surface friction which is detected by RCM 411 sensor in this study, and $\hat{Y}$ is the predicted road surface friction of the proposed prediction model. Typically, the MAE is used to measure the absolute error associated with a prediction, the MAPE presents a measure of the percentage of average misprediction of the model and the MSE measures the relative error for a prediction. The prediction model with the smaller values of MAE, MSE and MAPE performs better.

# 3   Data

## 3.1   Testing Field

The data used in this study were collected by an on-vehicle RCM 411 sensor from the road segments of E75 route from Sodankylä to Kemi in Finland. The total length of the road is 186 miles. In winter season, from October to next April, the air temperature is historically relatively low in this area. It could be minus 40 Celsius, and average minimum air temperature is about minus 15 Celsius. The average maximum air temperature is still under the ice point of water for the most of time. In the other seasons, the air temperature is not as high as in normal areas. The historical average maximum air temperature in July is about 20 Celsius. July is the month with the highest temperature in this area. Therefore, the study field definitely has road weather issues, like icing and snow happened most frequently. There was a detection vehicle equipped with an RCM 411 sensor kept collecting the road surface friction data since February 2017. In addition, the sensor system also estimates the road surface status based on road surface friction and other related measurements. Totally, 6 status are used to label the road surface, including Dry, Wet, Moist, Slush, Ice and

Snow or hoar frost. Figure 2 shows the distribution of calculated road surface status of two days which are random selected. One is in winter season and one is in summer season. In the figure, calculated road surface status was variated along the road for both tow selected days. The most part of the road was covered by snow or hoar frost in winter season, while the most part of the while the most part of the road was dried in summer season.

## 3.2   Data Description

### 3.2.1   Road Segmentation

As the road surface friction of a road segment has no spatial correlation with the surface condition of adjacent road segments, we proposed a spatial clustering method based on the K-means clustering algorithm to segment the study site into 1487 road segments. The detailed road segmentation method was introduced in the previous work [25]. Figure 2 visualizes the examples of road segmentation results in the small image patches. There is only one road surface status exists within the road segments. After segmenting the road based on the proposed criteria, the temporally average road surface friction could be used to represent the value of each road segment at a specific time stamp. Otherwise the value of the road surface friction could spatially vary a lot at one time stamp.

### 3.2.2   Statistical Description

The data used in this study was collected by Road Condition Monitor (RCM) 411 sensor, which is an optical sensing-based on-vehicle road surface condition sensor [8]. The dataset used in this study covers 446 days from February 17th, 2017 to May 9th, 2018. During this time period, the vehicle equipped

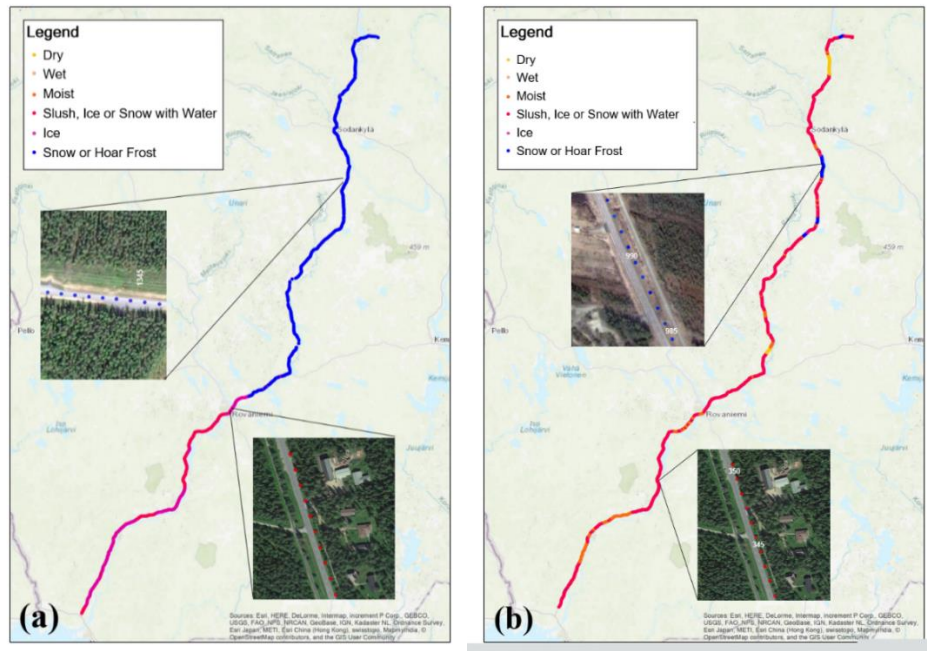

**Fig.2.** *Road Surface Status Distribution and Examples of Road Segmentation Results. (a) Winter Season, February 14th, 2018. (b) Summer Season, August 21st, 201*

with an RCM 411 sensor travelled through the testing field at least once per day, so that every road segment has at least one friction record for every single day.

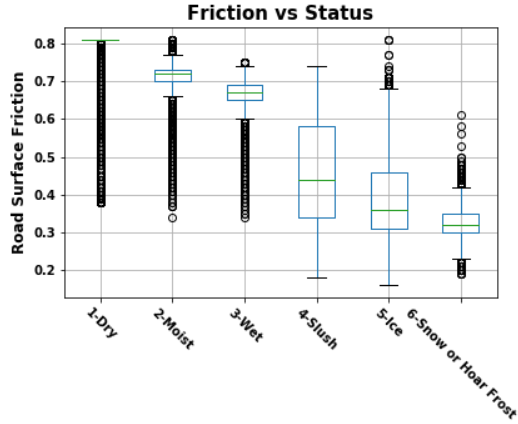

**Fig.3.** *Road Surface Friction Distribution of Different Surface Status in Study Site*

As mentioned before, besides the road surface friction, RCM 411 sensor also provide calculated road surface status with high accuracy [11]. In most cases, road surface status is defined or measured based on the friction information. In addition, the road surface friction coefficient is the most important indicator to characterize the anti-sliding performance of a road segment. Figure 3 shows the road surface friction distribution with different road surface status in the study site. As seen in the plot, the road surface friction decreased when the road surface status got worse. When the road surface was not covered by any type of ice or snow, the average road surface friction ranged from 0.75 to 0.8. Once any type of ice or snow existed on the road surface, the average road surface friction dropped under 0.5. According to the literature, the stopping distance is almost doubled when road surface friction drops from 0.8 to 0.5 [38].

In addition, the road surface friction also has time series features. Figure 4 presents the road surface friction of 5 selected road segments in the time period from February 17th, 2017 to May 9th, 2018. Basically, the road surface friction of the 5 selected share a similar time-series trend. During the time period from March to October 2017, the road surface friction was around 0.7 for the most of time. In other months, the road surface friction was fluctuated a lot, and low friction value happened more frequent due to the winter weather influence.

### 3.3 Data Preparation for GRU-D Networks and Baseline Models

Since it is assumed no spatial correlation between road segments, each road will be modelled separately. Thus, the spatial dimension of the input data of each road is set as $P = 1$. The unit of time-step for road surface friction detection is set as 1-day, then, the dataset has 446 time-steps for each road segment. Suppose the number of the time-lag is set as $T = t$ with 1 days between each time-step, which means the model used the data in previous $t$ consecutive time-steps to predict the road surface friction in the following $t + 1$ time-step. Based on the previous work, $T = 7$ was demonstrated to generated the best predictive performance in this application [25]. Then the dataset is separated into samples with $T = 7$ time-step, and the sample size is $N = 446 - 7 =$ 439. Thus, each sample of the input data, $X_n$, is a 2-dimensional vector with the dimension of $[T, P] = [7,1]$, and each sample of the output data is a 1-dimensional vector with 1 component. The input of the model for each road segment is a 3-dimensional vector which dimension is $[N, T, P] = [439, 7, 1]$ . Before feeding into the model, all samples are randomly divided into training set, validation set and test set with the ratio 7:2:1. All prediction models share the same data structure.

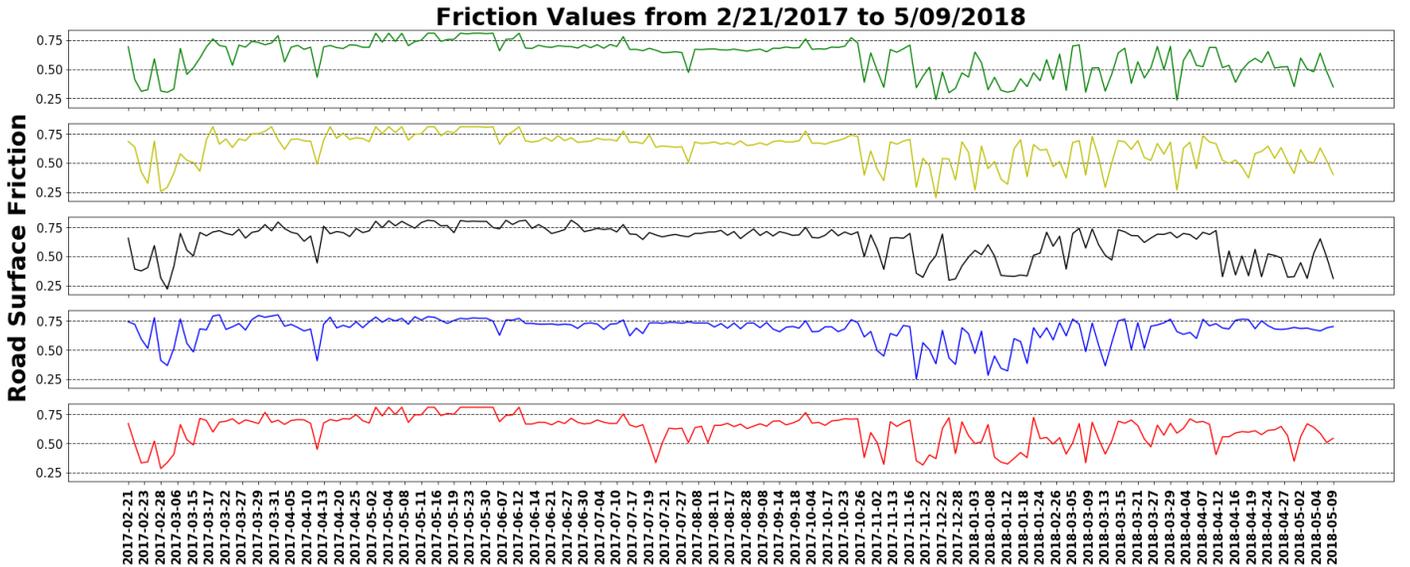

**Fig.4.** *Temporal Distribution of Road Surface Friction of 5 Selected Road Segments*

The original data set has no missing values. For testing the performance of GRU-D networks, the masking vectors were random generated by given a ratio of missing values. Then, the time interval vectors were calculated based on the corresponding masking vectors. As introduced in section 2.2.1, three simple methods were employed to evaluate the performance of the decay mechanism. The random generated missing values were imputed by these three baseline imputation methods before input to the prediction model. All prediction models were combined with baseline imputation methods for testing. The performance was compared with GRU-D networks and will be introduced in the next section.

## 4    Numerical Results

### 4.1    Prediction Performance Comparison of GRU and Baseline Prediction Models

The prediction models were tested using the dataset without any missing values. All prediction models were trained based on the same training data set for each road segment separately, and the predictive performance for each model was calculated using the testing dataset. Table 1 shows the predictive performance of all prediction models. The presented measurements were the averaged value of all road segments. In general, GRU networks outperformed all prediction models in terms of all three measurements. Among non-RNN prediction models, RF performed much better than SVR and Feed-Forward NN in terms of all three measurements, which makes sense due to the majority votes mechanism of RF model. The Feed-Forward NN had the worst predictive performance, which is caused by the sparsity of the data. The vanilla LSTM model performed better than all non-RNN models due to beneficial of long and short-term memory. The GRU networks achieved better performance than the LSTM model due to the sparse data features and less training parameters.

**Table 1** Predictive Performance of GRU Networks and Baseline Prediction Models without Missing Values

| Prediction Models | MSE | MAE | MAPE (%) |
|---|---|---|---|
| Feed-Forward NN | 0.0281 | 0.1482 | 26.81 |
| SVR | 0.0161 | 0.1111 | 19.99 |
| RF | 0.0113 | 0.0792 | 15.00 |
| LSTM | 0.0085 | 0.0697 | 12.85 |
| GRU | 0.0080 | 0.0640 | 11.92 |

**Table 2** Predictive Performance of Non-RNN Baseline Models with 80% Missing Values

| Non-RNN Models-Imputation Methods | MSE | MAE | MAPE (%) |
|---|---|---|---|
| Feed-Forward NN-Average | 0.0301 | 0.1539 | 28.52 |
| Feed-Forward NN-Last | 0.0301 | 0.1539 | 28.53 |
| Feed-Forward NN-Simple | 0.0301 | 0.1538 | 28.50 |
| SVR-Average | 0.0183 | 0.1152 | 21.01 |
| SVR-Last | 0.0175 | 0.1139 | 20.94 |
| SVR-Simple | 0.0192 | 0.1195 | 21.97 |
| RF-Average | 0.0128 | 0.0860 | 16.20 |
| RF-Last | 0.0119 | 0.0809 | 15.32 |
| RF-Simple | 0.0114 | 0.0820 | 15.71 |

**Table 3** Predictive Performance of and GRU-D and RNN Baseline Models with 80% Missing Values

| RNN Models-Imputation Methods | MSE | MAE | MAPE (%) |
|---|---|---|---|
| LSTM-Average | 0.0106 | 0.0794 | 15.84 |
| LSTM-Last | 0.0108 | 0.0808 | 15.46 |
| LSTM-Simple | 0.0105 | 0.0802 | 15.51 |
| GRU-Average | 0.0107 | 0.0811 | 15.80 |
| GRU-Last | 0.0108 | 0.0808 | 15.46 |
| GRU-Simple | 0.0109 | 0.0799 | 15.50 |
| **GRU-D** | 0.0102 | 0.0783 | 14.61 |

### 4.2    Prediction Performance Comparison of GRU-D and Baseline Imputation Methods

In this section, the dataset with 80% missing values was utilized as the input of the prediction models. For baseline missing data imputation methods, the missing values were imputed by three simple methods before modelling. Table 2 and Table 3 shows the predictive performance of non-RNN models and RNN models with three simple missing value imputation methods, respectively. In general, the predictive performance kept the same order with the predictive performance using dataset without missing values, but all prediction models had a little decrease in prediction accuracy due to the impact of missing values. The Last data imputation method outperformed Average and Simple methods except for RF models, and the Simple method achieved better performance than Average method. For the GRU-D networks, since both short-term features and empirical average value were considered in the decay mechanism. the GRU-D networks outperformed all other baseline imputation methods in terms of all three predictive performance measurements.

### 4.3    Analysis the Impact of Missing Rate on Prediction Accuracy

In order to investigate the impact of missing rate on predictive performance of the proposed GRU-D networks, the input data sets with the missing rate from 0% to 50% were used for testing. The data of all 1487 road segments were tested with the missing rates. Figure 5 shows the boxplots of three predictive measurements. When the missing rate equals 0%, the predictive performance of GRU-D networks has the same prediction accuracy with GRU networks. As the missing rate increased, the prediction precision gradually dropped. When the missing rate equals 50%, the GRU-D generated averaged 0.089 MAE with 0.0128 standard error. It is noticed that stopping distance and driver reaction time have a nonlinearly negative correlation curve with road surface friction. When the road surface friction lower than 0.5, the incremental stopping distance with one unit decrease in road surface friction is larger than the road surface friction larger than 0.5. 0.1 error in road surface friction could cause tens of meters error in stopping distance [38]. Thus, even the proposed GRU-D networks is capable to handle the missing values, the impact becomes more considerable in this implementation when the missing rate getting larger.

### 4.4    Learning Efficiency and Interpreting the Learned Decays

In this section, the learning efficiency and the learned input decays rate are analyzed. Figure 6 shows the validation loss curves versus the training epoch. Due to the early stopping mechanism is used in the training process, the numbers of training epochs are different. The GRU-D networks needs less epochs to converge than the GRU networks with other baseline imputation methods. In addition, the loss of the GRU-D networks decreased fastest and achieved lower loss among the compared models.

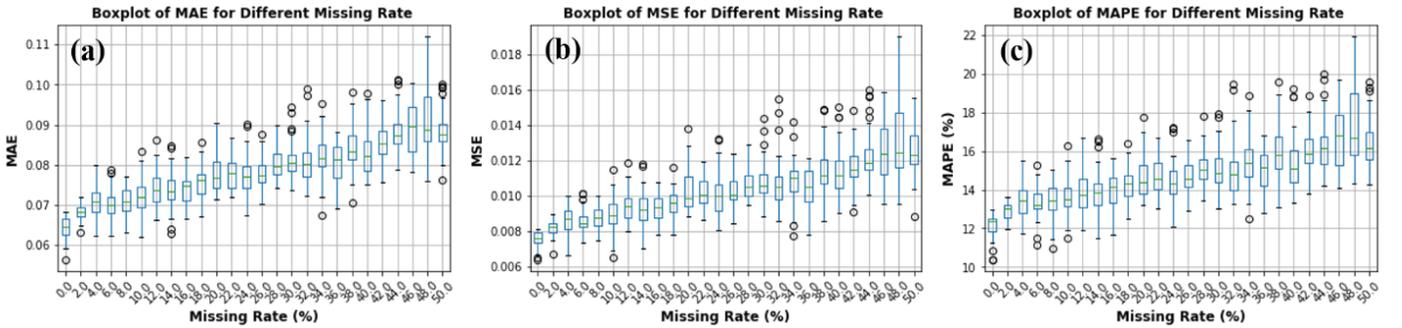

**Fig.5.** *Predictive Performance of GRU-D Networks with Different Missing Rates. (a) MAE. (b) MSE. (c) MAPE.*

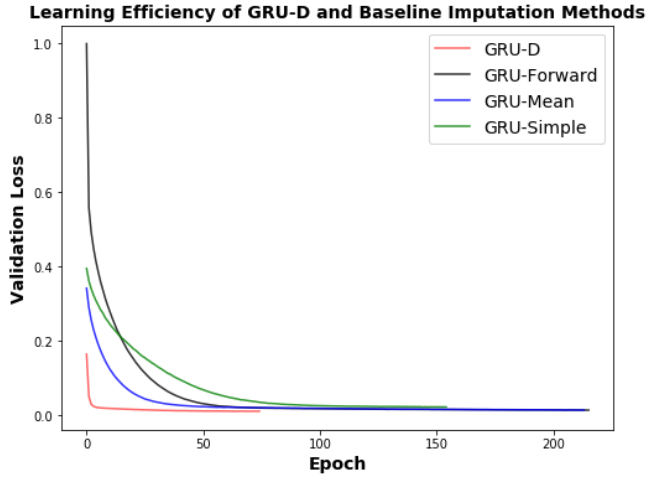

**Fig.6.** *Learning Efficiency of GRU Networks with Different Methods for Handling Missing Values*

Figure 7 shows the learned input decay rate $\gamma_{x_t}$, which can demonstrate how the informative missing pattern was utilized for missing value imputation. According to the figure, the learned input decay rate ranged from 0.8 to around 0.2. The decay rate dropped with a large slope when the time interval is smaller than 10 days. Then, the declining rate of the input decay rate decreased. Finally, it kept a constant when the time interval is larger than 20 days. This indicates that the value of the observation at the current and nearby time step is very important for the road surface friction prediction, and the model relies less on the previous observations with the time interval larger than 20 days.

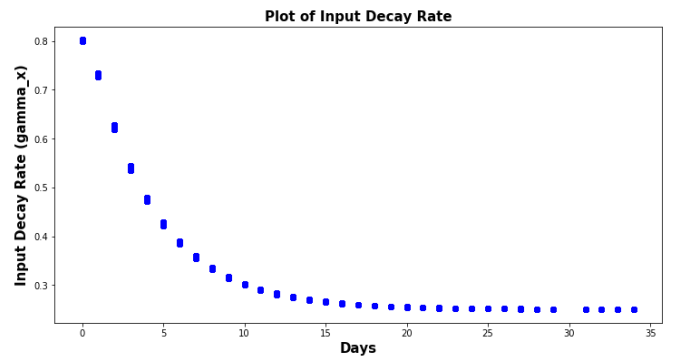

**Fig.7.** *Learned Input Decay Rate (x-axis is time interval $\delta_t^d$ from 0 35 days, y-axis is the value of learned input decay rates $\gamma_{x_t}^d$)*

## 5   Conclusion and Future Work

This study proposed a GRU-D networks for predicting road surface friction with missing values. The GRU-D unit handled the missing values by integrating a decay mechanism with the traditional GRU unit. Masking and time interval missing patterns are both considered in the GRU-D networks. To evaluate the predictive performance of the proposed method, Feed-Forward NN, SVR, RF and LSTM were selected as baseline prediction models, and three simple methods for data imputing were selected as baseline imputation methods. The evaluation results indicate the proposed model outperformed all baseline models with the most efficient learning efficiency. The analysis of learned decay rates presents the value of the observation at the current and nearby time step is more important for the road surface friction prediction at testing field, and the model relies less on the previous observations with the time interval larger than 20 days. For the future work, the model will be improved to increase the predictive accuracy and interpretability of the model parameters.

## 6   Acknowledgements

This research was supported by the multi-institutional project (Exploring Weather-related Connected Vehicle Applications for Improved Winter Travel in Pacific Northwest) of Pacific Northwest Transportation Consortium (PacTrans) USDOT University Transportation Center for Federal Region 10.